\pdfoutput=1
\documentclass[11pt]{article}
\usepackage[table]{xcolor}
\usepackage{acl}

\usepackage{times}
\usepackage{color}
\usepackage{latexsym}
\usepackage{hyperref}
\usepackage{booktabs}
\usepackage{multirow}
\usepackage{multicol}
\usepackage{marvosym}
\usepackage[T1]{fontenc}
\usepackage[utf8]{inputenc}
\newcommand{\nop}[1]{}
\usepackage{microtype}
\usepackage{setspace}

\usepackage{inconsolata}
\usepackage{graphicx}
\usepackage{arydshln}
\title{\emph{\LARGE{Reason from Fallacy}}: Enhancing Large Language Models' \\ Logical Reasoning through Logical Fallacy Understanding}
\author{Yanda Li$^\dag$, Dixuan Wang$^\dag$, Jiaqing Liang$^{\dag\ddag\textrm{\Letter}}$, Guochao Jiang$^\dag$, Qianyu He$^\dag$, \\\textbf{Yanghua Xiao$^{\dag\ddag}$, Deqing Yang$^{\dag\ddag\textrm{\Letter}}$ }\\
$^\dag$School of Data Science, Fudan University, Shanghai, China\\
$^\ddag$Shanghai Key Laboratory of Data Science, Shanghai, China\\
$^\dag$\texttt{\{ydli22, dxwang23, gcjiang22, qyhe21\}@m.fudan.edu.cn} \\
$^\ddag$\texttt{\{liangjiaqing, shawyh, yangdeqing\}@fudan.edu.cn}\\
}
\begin{document}
\maketitle
\begin{abstract}
Large Language Models (LLMs) have demonstrated good performance in many reasoning tasks, but they still struggle with some complicated reasoning tasks including logical reasoning. One non-negligible reason for LLMs' suboptimal performance on logical reasoning is their overlooking of understanding logical fallacies correctly. To evaluate LLMs' capability of logical fallacy understanding (LFU), we propose five concrete tasks from three cognitive dimensions of WHAT, WHY, and HOW in this paper. Towards these LFU tasks, we have successfully constructed a new dataset LFUD based on GPT-4 accompanied by a little human effort. Our extensive experiments justify that our LFUD can be used not only to evaluate LLMs' LFU capability, but also to fine-tune LLMs to obtain significantly enhanced performance on logical reasoning. 
%Our findings suggest that understanding logical fallacies can help to better avoid them, thereby significantly enhancing LLMs' overall logical reasoning ability.
\end{abstract}

\section{Introduction}

As a cognitive process, \emph{logical reasoning} plays an important role in many intellectual activities, such as problem solving, decision making and planning~\cite{huang2022towards}. Up to now, a lot of efforts have been dedicated to logical reasoning based on language models~\cite{Cresswell1973-CRELAL-3, kowalski1974logic, iwanska1993logical, liu2020logiqa}. More recently, the popularity of large language models (LLMs) such as ChatGPT ~\cite{ouyang2022training} and GPT-4~\cite{openai2023gpt} stimulates the growth of research on LLM-based logical reasoning. Compared to traditional small language models, LLMs have demonstrated better performance in many reasoning tasks.
% ~\cite{wei2022emergent}.

\begin{figure}
    \centering
    \includegraphics[width=1.05\linewidth]{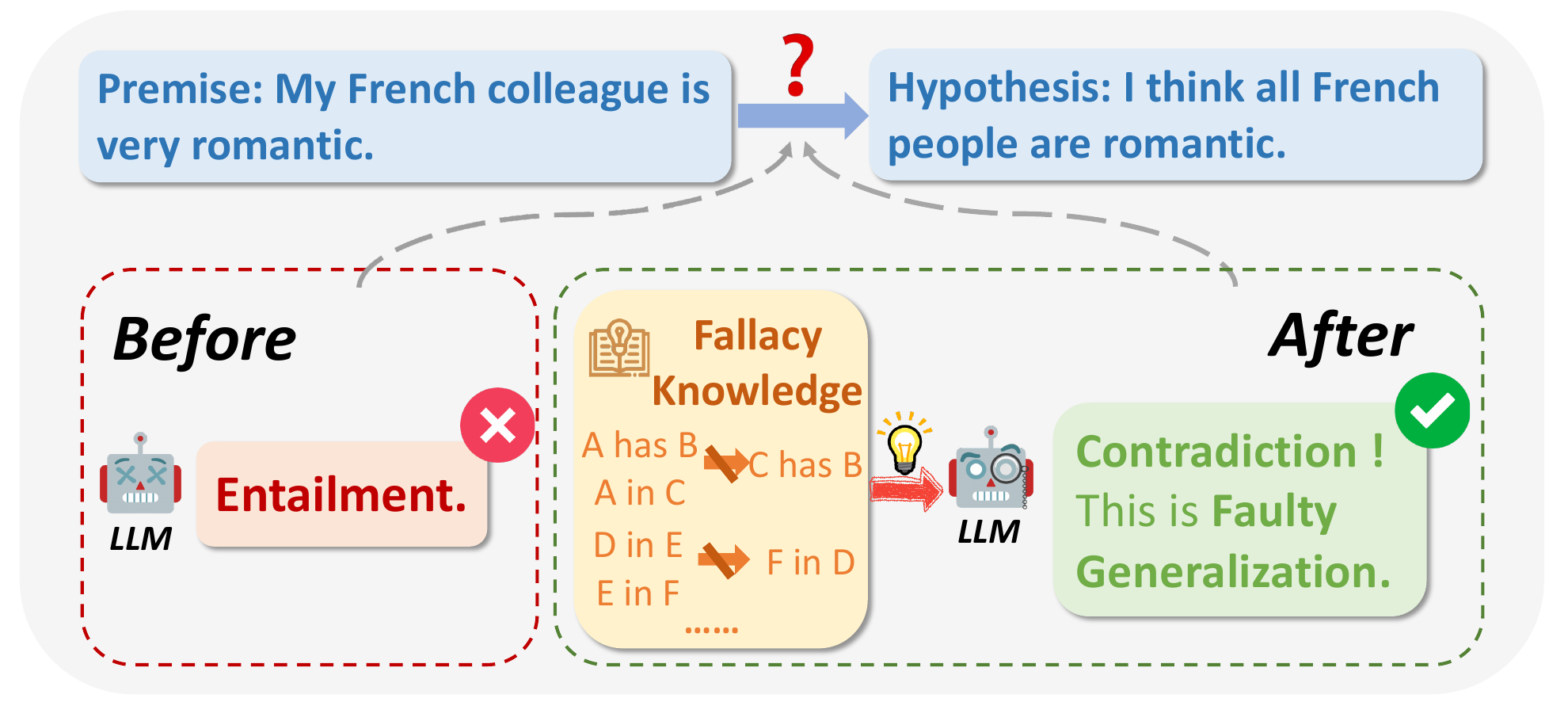}
    \caption{%\textbf{Getting knowledge from logical fallacies.} 
    LLMs have deficiencies in logical reasoning. Once they understand logical fallacies, they know how to avoid logical fallacies, and thus improve their performance in various logical reasoning tasks.}
    \label{fig:learn}
    %\vspace{-0.3cm}
\end{figure}

However, LLMs still struggle with some more complex reasoning tasks including logical reasoning. One non-negligible reason for LLMs' suboptimal performance on logical reasoning is their overlooking of understanding \emph{logical fallacies} correctly. As early as 350 BC, Aristotle first proposed the concept of logical fallacy in his work \emph{Sophistical Refutations} \cite{aristotle2006sophistical}. Since then, logical fallacies have gradually become an important issue that should be noticed in our lives. ``\textit{Thou shalt not commit logical fallacies!}'' has even become a worldwide popular idiom to remind us not to commit logical fallacies. By definition, logical fallacies refer to the errors in reasoning~\cite{tindale2007fallacies}, and they usually happen when the premises are not relevant or sufficient to draw the conclusions. Many previous works ~\cite{liu2020logiqa, yu2020reclor, joshi2020taxinli, han2022folio} have focused on evaluating LLM logical reasoning capabilities from the perspective of deductive reasoning, natural language inference, reading comprehension, etc. However, few works focus on logical fallacies, which is in fact the major reason causing logical inconsistency in the sentences.

\citeauthor{chen2023learning} have observed that, LLMs often commit
%stumble across 
logical fallacies in logical reasoning, such as \textit{"Either protect the environment or develop the economy."} (false dilemma) and \textit{"Some roses are not red because not all roses are red."} (circular reasoning). 
It has been found that language models could avoid mistakes only when they understand what mistakes are \cite{chen2023gaining, an2023learning}, 
which justifies the ancient Greek philosopher Epicurus's saying ``\textit{The mistake is the first step to save yourself.}'' 
Based on our empirical studies, we have also found that the logical reasoning capability of LLMs is closely related to their understanding of logical fallacies.

The previous studies related to logical fallacy~\cite{jin-etal-2022-logical, sourati2023case, srivastava2023beyond} only focus on logical fallacy detection, %but their findings cannot fully illustrate LLMs' capability to understand fallacies. These studies~\cite{jin-etal-2022-logical, sourati2023case, srivastava2023beyond} 
i.e., the identification and classification of logical fallacies, rather than systematically evaluating LLMs' capability of logical fallacy understanding (LFU), not to mention improving LLMs' LFU capability. Moreover, they have not explored the relationships between LFU and logical reasoning, which is crucial to improve LLMs' capability of logical reasoning through enhancing their LFU capability.
To address this problem, we focus on evaluating and enhancing LLMs' LFU capability in this paper, so as to enhance their capability of logical reasoning.

Nonetheless, our work has to face several challenges as follows. First, we need to formalize the concrete tasks for LFU, since no previous studies focus on this problem. Second, we need a new dataset specific to LFU, as the previous datasets of logical fallacies \cite{jin-etal-2022-logical} only contain the logical fallacy types presenting in the sentences. To this end, we should propose a framework of constructing the LFU dataset towards the concrete LFU tasks, and then truthfully evaluating LLMs' LFU capability with the dataset.
%as no previous studies focus on the evaluation of LLMs' logical fallacy understanding. Furthermore, there is a lack of data sources, so we need to build the datasets with synthetic data.

To overcome these challenges, we primarily focus on constructing a dataset for LFU in this paper, of which the samples are generated to evaluate models' achievement on the following five LFU tasks corresponding to three cognitive dimensions of \textbf{WHAT}, \textbf{WHY}, and \textbf{HOW}~\cite{swanborn2010case}.

\begin{enumerate}
    \setlength{\itemsep}{0pt} % 定义列表项之间的距离
    \setlength{\parsep}{0pt}  % 定义段落之间的距离
    \setlength{\parskip}{0pt} % 定义列表和其他元素之间的距离
    \item \textbf{WHAT}-Identification (Task 1) and Classification (Task 2): identifying whether the given sentence contains a logical fallacy and which type of logical fallacy it is.
    \item \textbf{WHY}-Deduction (Task 3) and Backward Deduction (Task 4): capturing the reasons causing the logical fallacy in the sentence. 
    \item \textbf{HOW}-Modification (Task 5): correcting the logical fallacy in the sentence.
\end{enumerate}

Our proposed LFU tasks simulate the human understanding process of logical fallacies. Towards these tasks, we design a pipeline framework to automatically generate and synthesize a high-quality dataset, namely \textbf{Logical Fallacy Understanding Dataset (LFUD)}, based on GPT-4 accompanied by a little human effort.
Specifically, we first collect some sentences
as the \emph{propositions} (statements) which are the basic logic units and used to generate the sentences containing logical fallacies. Then, with the help of GPT-4, we generate sentences based on the propositions with twelve typical logical fallacy types \cite{jin-etal-2022-logical}. And for each LFU task we propose, the instances of each fallacy type are synthesized. 
Then, we use our LFUD to evaluate the LFU capability of some representative LLMs. For the ultimate objective of our work, i.e., enhancing LLMs' capability of logical reasoning, we further fine-tune these LLMs with the instances in LFUD. Our extensive experiments reveal that fine-tuning LLMs with LFUD can significantly enhance their logical reasoning capability. %This effectively validates the value of our work presented in this paper.

In summary, our main contributions in this paper include:
    
1. Inspired by the three cognitive dimensions of \textbf{WHAT}, \textbf{WHY}, and \textbf{HOW}, we propose five concrete tasks which can truthfully evaluate LLMs' performance on LFU.

2. Towards our proposed five LFU tasks, we devise a new framework for constructing a high-quality dataset, namely LFUD, to evaluate LLMs' LFU capability, so as to enhance LLMs' performance on logical reasoning.

3. The LFUD we constructed includes 4,020 instances involving 12 logical fallacy types. Our extensive experiments have demonstrated that our LFUD can not only evaluate LLMs' LFU capability, but also improve LLMs' capability of logical reasoning through fine-tuning LLMs with LFUD samples in terms of the LFU tasks.

\section{Related Work}
\begin{figure*}
    \centering
    \includegraphics[width=1.02\linewidth]{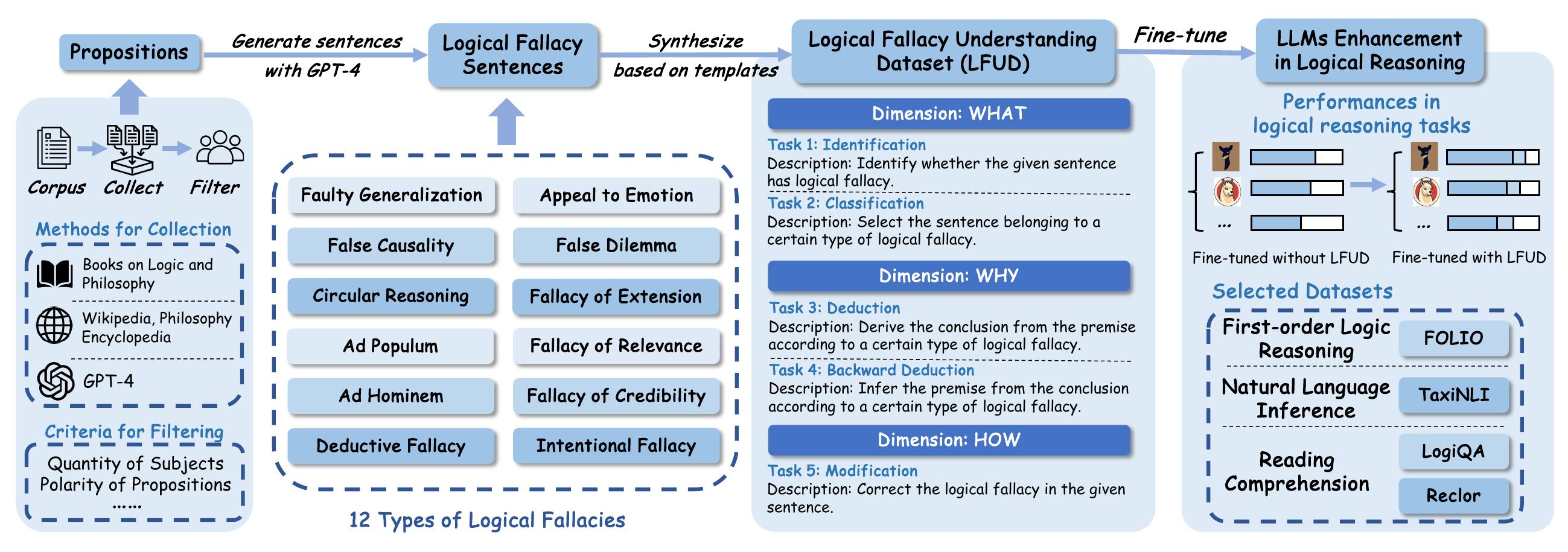}
    \vspace{-0.4cm}
    \caption{Our framework of constructing LFUD and fine-tuning LLMs with LFUD to enhance logical reasoning. At first, we collected some propositions, based on which the sentences with the logical fallacies of 12 types were generated by GPT-4. Then, for the five LFU tasks we proposed, the QA instances were synthesized based on the previous generated sentences. Finally, we fine-tuned LLMs with LFUD, revealing that fine-tuning LLMs with LFUD can significantly enhance their logical reasoning capability.}
    \label{fig:pipeline2}
    %\vspace{-0.2cm}
\end{figure*}

\paragraph{Logical Reasoning}

Up to now, a lot of efforts have been dedicated to logical reasoning based on language models~\cite{Cresswell1973-CRELAL-3, kowalski1974logic, iwanska1993logical, liu2020logiqa}. In particular, how to evaluate the models’ logical reasoning capability has attracted increasing attention, including deductive reasoning~\cite{ontanon2022logicinference, han2022folio}, natural language inference (NLI)~\cite{yanaka2019help, joshi2020taxinli, liu2021natural} and multi-choice reading comprehension (MRC)~\cite{liu2020logiqa, yu2020reclor, wang2022lsat}. Recently, the power of LLMs has stimulated the research on logical reasoning with LLMs, including LLMs evaluation~\cite{yu2023nature, blair2023can, teng2023glore}, and LLMs enhancement~\cite{zhang2023improved, chen2023learning}. Despite these works' achievements, enhancing LLMs' logical reasoning capability remains a non-negligible challenge. The major reason for logical inconsistencies in many sentences is the misunderstanding of logical fallacies, which is still under-explored in the research field of logic. 

\paragraph{Logical Fallacy}
Logical fallacy is the main reason for the logical inconsistencies presenting in our life. As early as 350 BC, Aristotle first proposed the concept of logical fallacy in his work \emph{Sophistical Refutations} \cite{aristotle2006sophistical}. Since then, logical fallacies have gradually gained attention in human society. In recent years, the studies related to logical fallacies mainly focused on dataset construction~\cite{habernal-etal-2018-name, martino2020semeval, jin-etal-2022-logical} and fallacy classification~\cite{stab-gurevych-2017-recognizing, goffredo2022fallacious, jin-etal-2022-logical, payandeh2023susceptible}. For instance, \citeauthor{jin-etal-2022-logical} first proposed the task of Logical Fallacy Detection, presenting a framework of 13 logical fallacy types, and evaluated all sentence samples on a classification task. \citeauthor{sourati2023case} proposed a Case-Based Reasoning method that classifies new cases of logical fallacy by language-modeling-driven retrieval and the adaptation of historical cases. However, there is no work to systematically evaluate LLMs' capability of logical fallacy understanding (LFU). For the first time, our work in this paper proposes a new dataset specific to LFU represented by five concrete tasks corresponding to three cognitive dimensions of \textbf{WHAT}, \textbf{WHY}, and \textbf{HOW}.

\paragraph{Learning from Synthetic Data}
Synthesizing data for model training has gradually gained popularity along with the advancements of language models. This approach is particularly beneficial for tasks that are difficult to be constructed or those with scarce data resources~\cite{moller2023prompt}. Currently, synthetic data has been applied in various tasks such as relation extraction~\cite{papanikolaou2020dare}, text classification~\cite{chung2023increasing}, irony detection~\cite{abaskohi-etal-2022-utnlp}, translation~\cite{sennrich2015improving}, and sentiment analysis~\cite{maqsud2015synthetic}. For example, \citeauthor{josifoski2023exploiting} proposed a strategy to design an effective synthetic data generation pipeline and applied it to closed information extraction. In addition, \citeauthor{li2023synthetic} conducted a series of experiments to evaluate the effectiveness of LLMs in generating synthetic data to support model training for different text classification tasks. Beyond these fundamental tasks, \citeauthor{eldan2023tinystories} proposed to use LLMs with synthetic data to generate short stories typically for 3 to 4-year-old only containing words. But they did not focus on logical fallacy. We are the first to focus on the data augmentation strategies in LFU.

\section{Methodology of Dataset Construction}\label{section3}
In this section, we present the pipeline of constructing our LFUD, of which the overall framework is depicted in Figure~\ref{fig:pipeline2}. Starting from the propositions, we detail the steps of synthesizing the samples towards five LFU tasks and the twelve representative logical fallacy types.

\nop{
\subsection{Process Overview of Data Generation}

Starting from the proposition (statement), we generate our task data based on three dimensions. Given the inherent complexity of the tasks, it would be excessively costly to manually construct data, which means that we must build an automatic generation framework. To validate the capability of GPT-4 in logical fallacy understanding, we tested a collection of 200 sentences from the Big-Bench~\cite{srivastava2023beyond}, which consists of both correct and incorrect sentences. The results showed that GPT-4 has the ability to accurately identifying over 90\% of the instances whether they have logical fallacies, indicating its capability to identify logical fallacies and perform certain tasks related to logical fallacies. However, the capability of GPT-4 in directly creating complex tasks is limited. So we choose to start with propositions, the most basic unit of logical reasoning, to construct our tasks partially leveraging the capacities of GPT-4.

Firstly, we collect and filter N propositions from multiple sources. Subsequently, these propositions are utilized to generate fallacy sentences as an integral part of the inference chain with the power of GPT-4. In consideration of the existing thirteen types of logical fallacies~\cite{jin-etal-2022-logical}, a single proposition can lead to the generation of thirteen sentences with different types of logical fallacies. Then we synthetize tasks in three dimensions by combining logical fallacy sentences with different structures.
}

\begin{table}[t]
    \centering
    \resizebox{0.45\textwidth}{!}
    {
    \begin{tabular}{lr}
    \toprule
         \textbf{Statistic} &\textbf{Number} \\
         \midrule
         Singular Proposition & 54 \\
         Particular Proposition & 5\\
         Universal Proposition & 8\\
         \midrule
         Affirmative Proposition & 56\\
         Negative Proposition & 11 \\
         \midrule
         Propositions without Pronouns & 58 \\
         Propositions with Pronouns & 9 \\
         \midrule
         Propositions with Human Subjects & 52 \\
         Propositions with Non-Human Subjects & 15 \\
         \bottomrule
    \end{tabular}}
    \vspace{-0.2cm}
    \caption{Some statistics of the 67 propositions in LFUD.}
    \label{tab:proposition}
    \vspace{-0.2cm}
\end{table}

\subsection{Acquiring Propositions}

At the first step of constructing our LFUD, we collected some propositions which were subsequently used for generating the sentences presenting various logical fallacies. According to \citet{Hurley2000-HURACI}, a proposition is one sentence that is either true or false. We considered several sources of proposition collection, including some authoritative books of logic and philosophy~\cite{Hurley2000-HURACI, Hausman2012-HAULAP}, open websites such as Wikipedia and \href{https://plato.stanford.edu/index.html}{Stanford Encyclopedia of Philosophy}. 
In addition, LLMs can be utilized to generate some propositions for enriching proposition diversity. 
To seek the satisfactory LLM for generating propositions, we tested some representative LLMs' identification performance on 200 instances from the Big-Bench~\cite{srivastava2023beyond}, consisting of correct and incorrect (logical fallacy) sentences. The results showed that GPT-4 can correctly identify in over 90\% of the sentences whether they have logical fallacies, despite the limited capability in directly generating complex tasks. Thus, we leveraged GPT-4 to generate more propositions and subsequent sentences presenting logical fallacies.

The considerable propositions should be simple and intuitive, but diverse. Finally, we filtered out 67 propositions and the relevant statistics are listed in Table~\ref{tab:proposition}. The following sentences are the proposition examples:

    \noindent 1. Everyone in my family has never been to Europe.
    
    \noindent 2. X accepted Y's suggestion.
    
    \noindent 3. Michael had dinner at an Italian restaurant.

\begin{table}[t]
    \centering
    \resizebox{0.5\textwidth}{!}
    {
    \begin{tabular}{p{0.5\textwidth}}
    \toprule
    \begin{minipage}{\linewidth}
        \setstretch{0.9} % 设置行间距
        \textcolor{gray}{/* }\textcolor{gray}{\textit{Generation Instruction }}\textcolor{gray}{*/}\\
            {\fontsize{10pt}{12pt}\selectfont As a logician, when presented with a proposition, your objective is to simulate the way of human thinking, generating a sentence with specific type of logical fallacy. 
            The generation should follow these instructions:} \\
            {\fontsize{10pt}{12pt}\selectfont 1. Generate the sentence with \textbf{Faulty Generalization}. Faulty Generalization occurs when ... (Detailed description)}\\
            {\fontsize{10pt}{12pt}\selectfont 2. The sentence should have complete premise and conclusion, but try not to make it too long.}\\
        \textcolor{gray}{/* }\textcolor{gray}{\textit{Three demonstration examples }}\textcolor{gray}{*/}\\
            {\fontsize{10pt}{12pt}\selectfont \textbf{Proposition 1:} Neither of the classes I took at UF were interesting.}\\
            {\fontsize{10pt}{12pt}\selectfont \textbf{Result 1: } A college is not a good college if none of its classes are interesting. Neither of the classes I took at UF were interesting, so UF is not a good college.}\\
            \textellipsis\\
            %{\fontsize{10pt}{12pt}\selectfont \textbf{Proposition 3:}\textellipsis}\\
            %{\fontsize{10pt}{12pt}\selectfont \textbf{Result 3:}\textellipsis}\\
        \textcolor{gray}{/* }\textcolor{gray}{\textit{Input the proposition }}\textcolor{gray}{*/}\\
            {\fontsize{10pt}{12pt}\selectfont \textbf{Proposition:} Peter visited China last year.}\\
        \textcolor{gray}{/* }\textcolor{gray}{\textit{GPT-4's output }}\textcolor{gray}{*/}\\
            {\fontsize{10pt}{12pt}\selectfont  \textbf{Result:} \textcolor{green!50!black}{Peter visited China last year. Peter is a European. Therefore, all Europeans have been to China.}}
    \end{minipage}
    \\ 
    \bottomrule
    \end{tabular}}
    % \vspace{-0.4cm}
    \caption{A prompt case for GPT-4 to generate a sentence with the given logical fallacy type.}
    \label{tab:prompt}
    %\vspace{-0.4cm}
\end{table}

\subsection{Generating Sentences with GPT-4}
Given GPT-4's capability of natural language generation and logical fallacy identification, we directly used GPT-4 to generate the sentences presenting various logical fallacies in this step. To take into account the logical fallacies existing in our life as many as possible, we refered to the thirteen typical types of logical fallacies (as listed in Table \ref{tab:ablation3} and Appendix \ref{sec:appendix_fallacy_type}) proposed by~\citet{jin-etal-2022-logical}.

\begin{table*}[t]
  \centering
  \small
    \begin{tabular}{p{30pt} p{100pt} p{295pt} }
    \toprule
    \textbf{Dimension} & \textbf{Task name} & \textbf{Task definition}\\
    \midrule
    \multirow{2}{*}{\textbf{WHAT}} & Task1: Identification & Identify whether the given sentence has logical fallacy. \\        
    & Task2: Classification & Select the sentence belonging to a certain type of logical fallacy.  \\
    \midrule
    \multirow{2}{*}{\textbf{WHY}} & Task3: Deduction & Derive the conclusion from the premise according to a certain type of logical fallacy. \\
    & Task4: Backward Deduction & Infer the premise from the conclusion according to a certain type of logical fallacy. \\
    \midrule
    \textbf{HOW}   & Task5: Modification & Correct the logical fallacy in the given sentence. \\
    \bottomrule
    \end{tabular}%
    \vspace{-0.2cm}
    \caption{Five LFU tasks corresponding to three cognitive dimensions.}\label{tb:task}
  \label{tab:addlabel}%
  \vspace{-0.2cm}
\end{table*}%

Given a proposition and a certain logical fallacy type, we asked GPT-4 to generate a sentence of this logical fallacy type with a prompt, which contains the generation instruction and a demonstration example of the given logical fallacy type. Table \ref{tab:prompt} illustrates the prompt for GPT-4 about the type of Faulty Generalization.

Specifically, due to the rather vague definition of Equivocation provided by \citet{jin-etal-2022-logical}, and the scarcity of such fallacy instances in real life, GPT-4 can hardly understand Equivocation and generate corresponding sentences correctly. To ensure the quality of the sentences generated by GPT-4, we neglected Equivocation fallacy type and generated the sentences for the rest twelve logical fallacy types.

To ensure that the generated sentences meet the requirements, we further manually proofread the sentences with logical fallacies generated by GPT-4. Each generated sentence was proofread with two main areas of concern: structural integrity and validity of fallacies, as described in Appendix \ref{sec:appendix_crowdsourcing}, to ensure that the sentences made sense and met the requirements of specific fallacy type.
For each of the 67 propositions, we generated 12 sentences with GPT-4, each of which presents one logical fallacy type. 
Thus, we generated 804 sentences with logical fallacies in total. These sentences are used to synthesize the samples for concrete LFU tasks as follows.

\subsection{Proposing LFU Tasks and Synthesizing Task Instances}

To evaluate LLMs' capability of LFU, we need to design concrete evaluation tasks.
According to the principles of cognitive science \cite{swanborn2010case}, humans generally understand objects from three dimensions: \textbf{WHAT} it is, \textbf{WHY} it is, and \textbf{HOW} it operates, which are interconnected and progressive cognition levels. 
Inspired by these dimensions, we propose five concrete tasks which are used to verify models' capability of LFU. Table \ref{tb:task} lists the definitions of the five tasks.
Wherein, Task 1 and Task 2 belong to WHAT dimension, which identify whether the given sentence has the logical fallacy (of a certain type). Task 3 and Task 4 belong to WHY dimension, which verify whether the model captures the reason causing the logical fallacy in the sentence. The last Task 5 belongs to HOW dimension, which requires correcting the logical fallacy of the given type in the sentence. %With these three dimensions as the starting point, we have established five concrete tasks, with examples shown in Appendic~\ref{sec:appendix_taskintro}
Specifically, we synthesized multiple-choice questions for the first four tasks, and sentence generation questions for Task 5.
We further provided one toy example for each task in Appendix \ref{sec:appendix_taskintro}.

In fact, previous studies \cite{jin-etal-2022-logical, srivastava2023beyond} have focused on the two tasks of WHAT dimension, i.e., understanding what the logical fallacy in the sentence is. To the best of our knowledge, there are no studies concerning the tasks of {WHY} and {HOW} dimensions by now. But notably, the ultimate goal of LFU is to avoid logical fallacies, which requires us to understand the reasons causing logical fallacies and correct logical fallacies. Therefore, we paid more attention to the tasks of {WHY} and {HOW} dimension in this paper. %Next, we introduce how to generate the LFUD samples for these five tasks.
 
For each sentence with one of the twelve logical fallacy types generated in the previous step, we synthesized one QA instance for every LFU task with the question templates. 
%Every logical fallacy sentence corresponds to five task instances. Before synthesizing specific tasks, we initially establish templates for five tasks. 
For each LFU task, the question stems (without question options) of all instances are generated according to some templates, as shown in Appendix \ref{sec:appendix_taskintro}.
Particularly, for Task 3 and Task 4, we need to identify the premise and conclusion for the given sentence, and further provide question options. Thus, we directly asked GPT-4 to generate the results as we needed.

To minimize the impact of instruction design when asking LLMs to achieve these tasks, we first designed some candidate question templates to constitute a template pool in fact, and then randomly chose one template from the pool to generate the question for a certain LFU task. In addtion, we also shuffled the orders of question options. %Synthesizing the existing data on logical fallacy sentences with the templates, we get the data of five fallacy tasks we need, accomplish the dataset construction.
Finally, our LFUD contains 4,020 (QA) instances in total, involving 5 LFU tasks and 12 logical fallacy types, which stem from the 67 propositions and 804 sentences with logical fallacies.\footnote{LFUD is provided at \url{https://github.com/YandaGo/LFUD}}

\section{Evaluation}
\begin{table*}
\centering
\resizebox{0.98\textwidth}{!}{
\begin{tabular}{lccccccccccc}
\toprule
\multirow{2}{*}{\textbf{Datasets}} & \multirow{2}{*}{\textbf{FT Data}} & \multicolumn{2}{c}{\textbf{LLaMA2-13B}} & \multicolumn{2}{c}{\textbf{LLaMA2-7B}} & \multicolumn{2}{c}{\textbf{Vicuna-13B}} & \multicolumn{2}{c}{\textbf{Vicuna-7B}} & \multicolumn{2}{c}{\textbf{Orca2-7B}}\\
&  & \textbf{Acc.} & $\Delta\%$ & \textbf{Acc.} & $\Delta\%$  & \textbf{Acc.} & $\Delta\%$  &\textbf{Acc.} & $\Delta\%$  &\textbf{Acc.} & $\Delta\%$  \\
\toprule
\textbf{LogiQA2.0}  
% & LFUD & 37.98 & 36.79 & 43.96 & 41.60 & 47.07\\
                    & Origin & \underline{45.55} & - & \underline{42.30} & - & 52.74 & - & \underline{47.71} & - & \underline{54.39} & -\\
                    & Origin+LOGIC & 44.66 & -1.95 & 35.62 & -15.79 & \underline{53.37} & 1.19 & 45.10 & -5.47 & 52.93 & -2.68\\
                    % & Ori+LFUD(\#) & 46.06 & 41.54 & 52.16 & 46.82 & 51.97  \\
                    & Origin+LFUD & \textbf{47.90} & 5.16 & \textbf{43.13} & 1.96 & \textbf{55.85} & 5.90 & \textbf{47.84} & 0.27 & \textbf{56.55}& 3.97\\
\midrule
\textbf{Reclor}     
% & LFUD & 40.00 & 40.00 & 50.00 & 47.20 & 50.20\\
                    & Origin & \underline{47.20} & - & 40.40 & - & \underline{54.40} & - & \underline{49.20} & - & \underline{55.80} & - \\
                    & Origin+LOGIC & 46.20 & -2.12 & \underline{42.20} & 4.46 & 54.00 & -0.74 & 47.80 & -2.85 & \underline{55.80} & 0.00\\
                    % & Ori+LFUD(\#) & 43.00 & 40.80 & 52.60 & 49.40 & 53.60  \\
                    & Origin+LFUD & \textbf{50.20} & 6.36 & \textbf{46.40} & 14.85 & \textbf{57.00} & 4.78 & \textbf{51.80} & 5.28 & \textbf{58.20}& 4.30\\
\midrule
\textbf{TaxiNLI}    
% & LFUD & 37.22 & 37.07 & 40.65 & 35.78 & 67.30\\
                    & Origin & \underline{68.54} & - & \underline{62.68} & - & \underline{78.91} & - & \underline{77.47} & - & \underline{82.33} & - \\
                    & Origin+LOGIC & 40.60 & -40.76 & 58.80 & -6.19 & 77.92 & -1.25 & 76.18 & -1.67 & 82.18 & -0.18 \\
                    % & Ori+LFUD(\#) & 65.81 & 62.03 & 77.92 & 76.18 & 82.38  \\
                    & Origin+LFUD & \textbf{73.70} & 7.53 & \textbf{67.26} & 7.31 & \textbf{79.76} & 1.08 &\textbf{77.77} & 0.39 & \textbf{84.02}& 2.05 \\
\midrule
\textbf{FOLIO}      
% & LFUD & 39.22 & 38.73 & \underline{37.25} & 35.78 & 56.37 \\
                    & Origin & 61.76 & - & 50.98 & - & \underline{36.76} & - & \underline{50.49} & - & 72.55 & - \\
                    & Origin+LOGIC & \underline{62.25} & 0.79 & \underline{52.45} & 2.88 & 36.28 & -1.31 & 45.10 & -10.68 & \underline{73.53}  & 1.35\\
                    % & Ori+LFUD(\#) & - & - & - & - & -  \\
                    & Origin+LFUD & \textbf{66.18} & 7.16 & \textbf{59.31} & 16.34 & \textbf{44.61} & 21.35 & \textbf{56.37} & 11.65 & \textbf{76.47} & 5.40 \\
\bottomrule
\end{tabular}}
\caption{LLMs' accuracy(\%) on the four logical reasoning tasks (datasets) after being fine-tuned with different data. Origin represents fine-tuning the LLMs with the original training data in the logical reasoning datasets. 
$\Delta\%$ is accuracy improvement relative to Origin.  
The best accuracy scores are \textbf{bolded} and the second best scores are \underline{underlined}.}
\label{tab:finetune}
\end{table*}

\subsection{Experiment Setup}
\paragraph{Datasets} To evaluate LLMs' performance on logical reasoning, we used four representative datasets including FOLIO~\cite{han2022folio}, TaxiNLI~\cite{joshi2020taxinli}, LogiQA~\cite{liu2020logiqa}, and Reclor~\cite{yu2020reclor} in our experiments. 
%These four datasets essentially cover the core work in the field of logical reasoning. 

FOLIO focuses on first-order logic reasoning (FOL) that is a classical deductive reasoning task. TaxiNLI is specific to natural language inference (NLI) that tests the logical relationship between a premise and a hypothesis. LogiQA and Reclor are the multi-choice reading comprehension (MRC) datasets, which choose the most suitable answer corresponding to the given text, could better reflect comprehensive logical reasoning abilities. The instances of the four datasets are shown in Appendix \ref{sec:appendix_logicalreasongdataset}. In addition to the training data in above four datasets and our LFUD, we also used the logical fallacy data
LOGIC \cite{jin-etal-2022-logical} to fine-tune LLMs. LOGIC (including LOGIC-CLIMATE) contains thirteen types of logical fallacy sentences, as shown in Appendix \ref{sec:appendix_fallacy_type}.

\paragraph{LLMs} We selected five popular LLMs in our experiments, including LLaMA2-7B, LLaMA2-13B \cite{touvron2023llama}, Vicuna-7B, Vicuna-13B \cite{chiang2023vicuna} and Orca2-7B~\cite{mitra2023orca}. 
When fine-tuning these LLMs, we set the learning rate to 2.5e-5 and the batch size to 8. To ensure the robustness of our results, we repeated all experiments for three times and reported the average performance (accuracy) scores.

\paragraph{Dataset Split}

For the 4,020 synthesized instances in our LFUD, we randomly selected 3,000 instances (corresponding to 600 sentences with logical fallacies) as the training set and the remaining 1,020 instances (corresponding to 204 sentences with logical fallacies) as the test set.
Given the instances of Task 1--4 (choice questions) have fixed answers, we only used the training samples (2,500 instances) of Task 1--4 to fine-tune the five LLMs. 
And we directly used some test samples of Task 5 to evaluate LLMs' cross-task learning capability on LFU, as presented in Subsection~\ref{sec:EvalLFU}. To balance the labels of logical right and fallacy in Task 1 instances, we appended 500 logically correct sentences of Big-Bench \cite{srivastava2023beyond}, and thus collected 2,900 training samples in our LFUD in total.

\subsection{Effectiveness on Enhancing LLMs' Logical Reasoning}
\subsubsection{Overall Performance}
To justify the value of our LFUD instances on enhancing LLMs' logical reasoning capability, we merged LFUD training samples with the original training samples in the four logical reasoning datasets, denoted by Origin, to fine-tune LLMs. We compared such a fine-tuning method with the method of fine-tuning LLMs only with Origin. 
In addition, we also compared the method of fine-tuning LLMs with Origin and some samples in LOGIC \cite{jin-etal-2022-logical}, which have the same number as the training samples in LFUD. 

Table~\ref{tab:finetune} lists the accuracy(\%) scores of all five LLMs on the four logical reasoning tasks (datasets) which were fine-tuned with Origin, Origin+LOGIC and Origin+LFUD, respectively. And the performance improvements of Origin+LOGIC and Origin+LFUD relative to Orign are also listed.
Based on the results in this table, we have the following observations and analysis.

\noindent 1. Appending the training samples in our LFUD to Origin when fine-tuning LLMs significantly enhances their performance on all logical reasoning tasks. It shows that learning the LFU tasks we proposed is indeed helpful to improve LLMs' capability of various logical reasoning.

\noindent 2. Although the samples in LOGIC are also the sentences with various logical fallacies, Origin+LOGIC cannot obtain the significant performance improvements of logical reasoning. Even worse, it degrades LLMs' logical reasoning performance, compared with Origin in some cases. It implies that, unlike our LFU tasks from WHAT, WHY and HOW, only identifying the logical fallacy presented by the sentences in LOGIC cannot result in LLMs' really capability of LFU. In addition, The samples in LOGIC are raw and unclean, with some examples consisting of even fallacy questions and fallacy definitions.

\begin{figure}[t]
    \centering
    \includegraphics[width=0.85\linewidth]{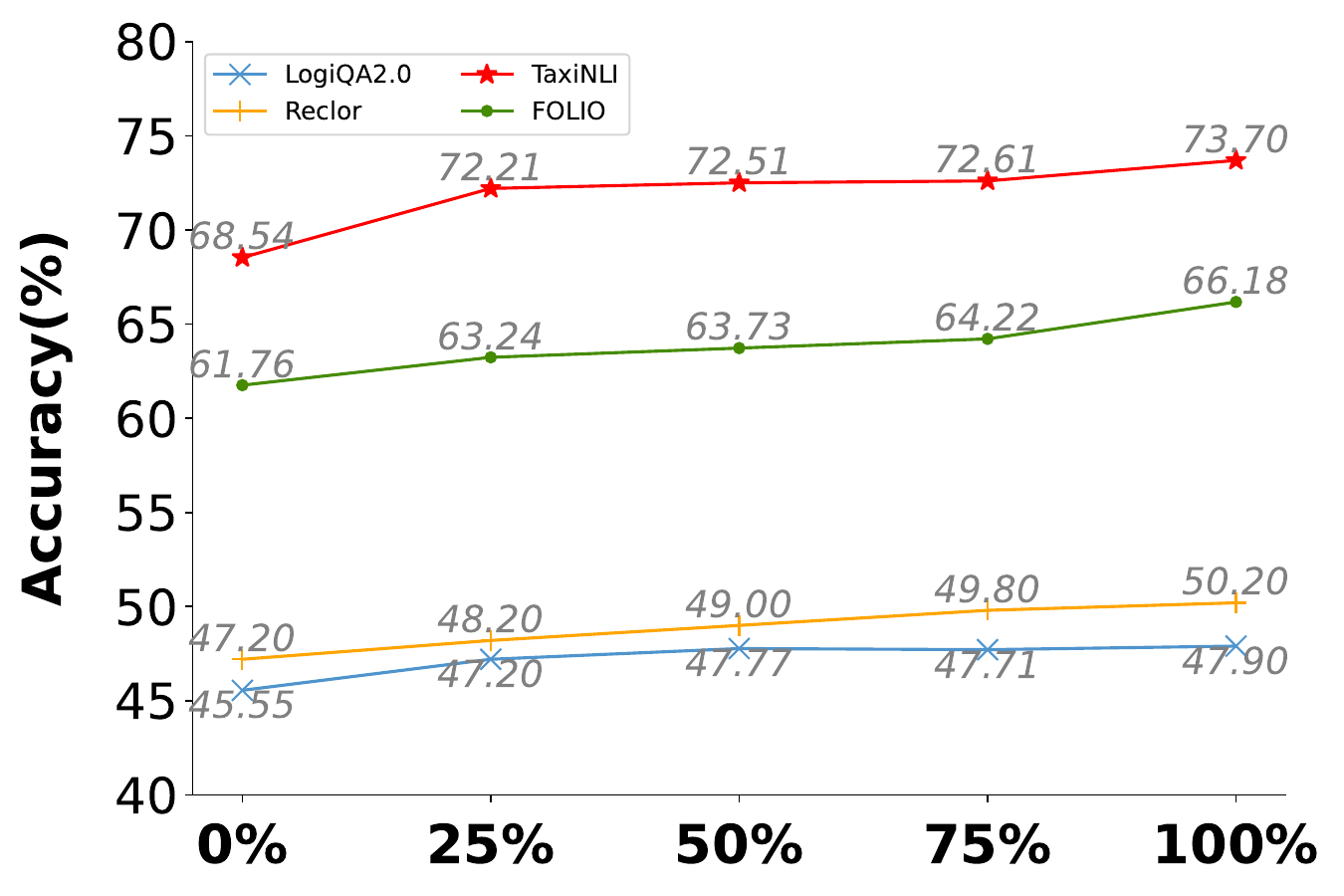}
    \vspace{-0.2cm}
    \caption{LLaMA2-13B's performance on the four logical reasoning tasks with different scales of LFUD training samples. %We fine-tuned LLaMA2-13B with the original training set mixed with fallacy data of LFUD ranging from 0\%, 25\%, 50\%, 75\% to 100\%.
    }
    \label{fig:Ablation1}
    
\end{figure}

\begin{table}[t]
\centering
\resizebox{0.48\textwidth}{!}{
\begin{tabular}{llllll}
\toprule
\multirow{2}{*}{\textbf{Models}} &  \multicolumn{2}{c}{\textbf{WHAT}} & \multicolumn{2}{c}{\textbf{WHY}} & \textbf{HOW}\\
&\textbf{Task1} & \textbf{Task2} & \textbf{Task3} & \textbf{Task4} & \textbf{Task5} \\
\midrule
\textbf{LLaMA2-7B} & 46.84 & 47.86 & 24.75 & 29.35 & 40.00\\
\textbf{LLaMA2-13B} & \underline{55.76} & 53.61 & 52.36 & 54.48 & 50.00\\
\textbf{Vicuna-7B} & 32.96 & 57.71 & 60.70 & 56.59 & 46.00\\
\textbf{Vicuna-13B} & 50.76 & 58.46 & 58.33 & 61.32 & 56.00\\
\textbf{ChatGPT} & 54.66 & \underline{73.88} & \underline{62.94} & \underline{70.65} & \underline{60.00}\\
\textbf{GPT-4} & \textbf{86.35} & \textbf{86.19} & \textbf{78.61} & \textbf{85.70} & \textbf{88.00}\\
\bottomrule
\end{tabular}}
\vspace{-0.2cm}
\caption{
Six representative LLMs' performance on our proposed five LFU tasks. To evaluate Task 5, we manually assessed LLMs' outputs for 50 randomly selected samples.}
\label{tab:result}
\end{table}

\subsubsection{Impacts of Different Factors in LFUD}
% 为了进一步证明我们数据集是有效的，我们使用LLaMA-13B，进行了几组消融实验。
To further validate LFUD's effectiveness on enhancing LLMs' logical reasoning capability, we also investigated the impacts of different factors in LFUD, including the scale of training data, LFU tasks and logical fallacy types. Due to space limitation, we only display the results of LLaMA2-13B.

\paragraph{Training Data Scale} To verify the impacts of training data scale, we respectively extracted 25\%, 50\%, and 75\% of the LFUD training data accompanied with Origin to fine-tune LLaMA2-13B, and then tested its performance on the four logical reasoning tasks. From Figure~\ref{fig:Ablation1} we can see that, LLaMA2-13B's performance improvement becomes more apparent as the training data scale increases, showing that even only a small part of LFUD samples is also valuable.

\paragraph{LFU Task} We fine-tuned LLaMA2-13B again with the training data excluding the instances of Task 1, Task 2, Task 3 and Task4, respectively. As shown in Table~\ref{tab:ablation2}, excluding any task's instances would lead to the performance decline of LLaMA2-13B. %As we can see, all four categories of tasks could enhance LLMs' logical reasoning ability.
% Comparatively, the impact of missing Task 1's instances is the more significant.

\begin{table}
\centering
\resizebox{0.5\textwidth}{!}{
\begin{tabular}{lcccc}
\toprule
\textbf{Task Category} & \textbf{LogiQA2.0} & \textbf{Reclor} & \textbf{TaxiNLI} & \textbf{FOLIO} \\
\midrule
\textbf{No Tasks} & 45.55 & 47.20 & 68.54 & 61.76 \\
\textbf{w/o Task1} & 46.69 & 49.80 & 69.53 & 63.73 \\
\textbf{w/o Task2} & 45.74 & 48.00 & 69.88 & 65.20 \\
\textbf{w/o Task3} & 47.46 & 49.00 & 72.01 & 64.22 \\
\textbf{w/o Task4} & 46.44 & 48.80 & 69.28 & 65.20 \\
\textbf{All Tasks} & \textbf{47.90} & \textbf{50.20} & \textbf{73.70} & \textbf{66.18} \\
\bottomrule
\end{tabular}}
\vspace{-0.2cm}
\caption{LLaMA2-13B's performance on the four logical reasoning tasks when excluding different LFU task's training instances.}
\label{tab:ablation2}
%\vspace{-0.3cm}
\end{table}

\paragraph{Logical Fallacy Type} Similarly, we respectively excluded the instances of each logical fallacy type from LFUD training data, and then tested LLaMA2-13B's performance. The results in Table \ref{tab:ablation3} indicate that every logical fallacy type contributes positively to LLM's logical reasoning capability.

\begin{table}
\centering
\resizebox{0.5\textwidth}{!}{
\begin{tabular}{lcccc}
\toprule
\textbf{Fallacy Type} & \textbf{LogiQA2.0} & \textbf{Reclor} & \textbf{TaxiNLI} & \textbf{FOLIO} \\
\midrule
\textbf{No Fallacy Data} & 45.55 & 47.20 & 68.54 & 61.76 \\
\textbf{w/o Faulty Generalization} & 46.56 & 49.80 & 71.91 & 64.71 \\
\textbf{w/o False Causality} & 46.69 & 47.60 & 72.56 & 62.75 \\
\textbf{w/o Circular Reasoning} & 46.12 & 49.80 & 72.95 & 64.22 \\
\textbf{w/o Ad Populum} & 46.25 & 47.60 & 72.85 & 64.22 \\
\textbf{w/o Ad hominem} & 46.95 & 48.60 & 69.53 & 65.20 \\
\textbf{w/o Deductive Fallacy} & 45.87 & 49.40 & 73.78 & 62.75 \\
\textbf{w/o Appeal to Emotion} & 47.65 & 49.80 & 69.93 & 63.73 \\
\textbf{w/o False Dilemma} & 46.12 & 50.00 & 73.10 & 63.24 \\
\textbf{w/o Fallacy of Extension} & 45.93 & 49.40 & 72.51 & 64.71 \\
\textbf{w/o Fallacy of Relevance} & 47.65 & 50.20 & 70.92 & 61.27 \\
\textbf{w/o Fallacy of Credibility} & 47.58 & 48.60 & 72.06 & 62.75 \\
\textbf{w/o Intentional Fallacy} & 46.88 & 49.80 & 69.48 & 65.69 \\

\textbf{All Fallacy Types} & \textbf{47.90} & \textbf{50.20} & \textbf{73.70} & \textbf{66.18} \\
\bottomrule
\end{tabular}}
% \caption{\textcolor{red}{Ablation results on Fallacy Type, in terms of accuracy(\%).}}
\vspace{-0.2cm}
\caption{LLaMA2-13B's performance on the four logical reasoning tasks  when excluding different logical fallacy type's training instances.}
\label{tab:ablation3}
\end{table}

\subsection{LFU Performance of LLMs}\label{sec:EvalLFU}
Next, we validate LLMs' capability of LFU through evaluating their performance on the LFU tasks. We want to investigate LLMs' inherent capability on LFU, thus we directly used all instances of each LFU task in LFUD as the test samples without fine-tuning them with the training data.

\paragraph{Performance on Each LFU Task}
Besides the previous four LLMs, we additionally considered ChatGPT \cite{ouyang2022training} and the latest GPT-4 \cite{openai2023gpt} (using OpenAI API with temperature 0.7) in LFU performance evaluation.
To balance the labels of Task 1, we added all 654 correct sentences in Big-Bench into Task 1's test data. Thus, we have a total of 1,458 instances for Task 1's evaluation. In addition, as Task 5 is to generate a new sentence rather than a fixed answer, we randomly selected 50 samples from its instances and manually assessed LLMs' outputs. 
All tested LLMs' performance is listed in Table \ref{tab:result}, showing that different LLMs' performance varies significantly on the five LFU tasks. 
%but overall, \textbf{understanding logical fallacies is a challenging task}. 
Among the LLMs, GPT-4 has much better performance than others on all tasks, justifying its strong capability of LFU. %However, other LLMs were less capable to understand logical fallacies to some extent. Among the LLMs, 
By contrast, LLaMA2-7B has the worst performance that is even worse than random selection.

\begin{table}[t]
    \centering
    \resizebox{0.3\textwidth}{!}{
    \begin{tabular}{lc}
    \toprule
         \textbf{Models} &\textbf{Accuracy(\%)} \\
         \midrule
         \textbf{LLaMA-2-7B} & 0.92 (6/654) \\
         \textbf{LLaMA-2-13B} & 1.99 (13/654)\\
         \textbf{Vicuna-7B} & 7.95 (52/654)\\
         \textbf{Vicuna-13B} & 26.61 (174/654)\\
         \textbf{ChatGPT} & 37.92 (248/654) \\
         \textbf{GPT-4} & 95.57 (625/654) \\
         \bottomrule
    \end{tabular}}
    \vspace{-0.2cm}
    \caption{LLMs' Performance on identifying 654 logically correct sentences of Task 1. %The results showed that LLMs except GPT-4 were likely to cater to the inquiries, admitting logical fallacies in sentences.
    }
    \label{tab:task1}
\end{table}

\begin{table}[t]
\centering
\resizebox{0.5\textwidth}{!}{
\begin{tabular}{lcccc}
\toprule
\textbf{Fallacy Type} & \textbf{Task 1} & \textbf{Task 2} & \textbf{Task 3} & \textbf{Task 4} \\
\midrule
\textbf{Faulty Generalization} & \underline{76.12} & \textbf{89.55} & 59.70 & 50.75\\
\textbf{False Causality} & 61.19 & 70.15 & \underline{67.16} & 65.67 \\
\textbf{Circular Reasoning} & 34.33 & 52.24 & 55.22 & 62.69 \\
\textbf{Ad Populum} & 65.67 & \underline{80.60} & \textbf{79.10} & \textbf{79.10} \\
\textbf{Ad hominem} & \textbf{77.61} & \textbf{89.55} & 59.70 & 59.70 \\
\textbf{Deductive Fallacy} & 40.30 & 49.25 & 62.69 & \underline{77.61} \\
\textbf{Appeal to Emotion} & 16.42 & 77.61 & 64.18 & \underline{77.61} \\
\textbf{False Dilemma} & 29.85 & 44.78 & 62.69 & 50.75 \\
\textbf{Fallacy of Extension} & 53.73 & 25.37 & 41.79 & 38.81 \\
\textbf{Fallacy of Relevance} & 25.37 & 37.31 & 11.94 & 44.78\\
\textbf{Fallacy of Credibility} & 40.30 & 53.73 & 61.19 & 64.18 \\
\textbf{Intentional Fallacy} & 68.66 & 31.34 & 47.76 & 64.18 \\
\bottomrule
\end{tabular}}
\vspace{-0.2cm}
\caption{Vicuna-13B's performance on Task 1--4 specific to each type of logical fallacies. The best accuracy scores are \textbf{bolded} and the second best scores are \underline{underlined}.}
\label{tab:fallacytype}
\end{table}

\paragraph{Identifying Logical Correctness} 
%Based on the results, we hold the idea that LLaMA2, Vicuna, and ChatGPT struggle with logical fallacies, but GPT-4 can do it. On one hand, we analyzed LLMs' performance on Task 1 with 654 correct instances, as shown in Table \ref{tab:task1}.
To further investigate whether LLMs really understand logical fallacies, we also asked LLMs to achieve Task 1 for the 654 sentences from Big-Bench that are logically correct (without logical fallacies). Their accuracy scores are listed in Table \ref{tab:task1}, showing that only GPT-4 has the satisfactory performance for this task. In fact, the rest LLMs tended to recognize the sentences as having logical fallacies for catering to Task 1's question. In addition, we also found these LLMs except for GPT-4 are easily influenced by the order of question options when achieving Task 1--4, indicating that they cannot well understand logical fallacies.

\paragraph{In Terms of Logical Fallacy Type} Besides, we evaluated LLMs' LFU performance on Task 1--4 in terms of a specific logical fallacy type. 
%to conducted a fine-grained analysis to check the performance of different fallacy types. The types of logical fallacies significantly impacted the performances across different tasks, which remained consistent for different LLMs. 
The results listed in Table \ref{tab:fallacytype} show that, LLMs exhibited better performance on the tasks of \textit{Faulty Generalization}, \textit{False Causality}, \textit{Ad populum} and \textit{Ad Hominem}. These four types of logical fallacies are more distinctive and more frequently present in our life, resulting in that LLMs have encountered more sentences with these logical fallacy types during their pre-training.

\begin{figure}
    \centering
    \includegraphics[width=0.9\linewidth]{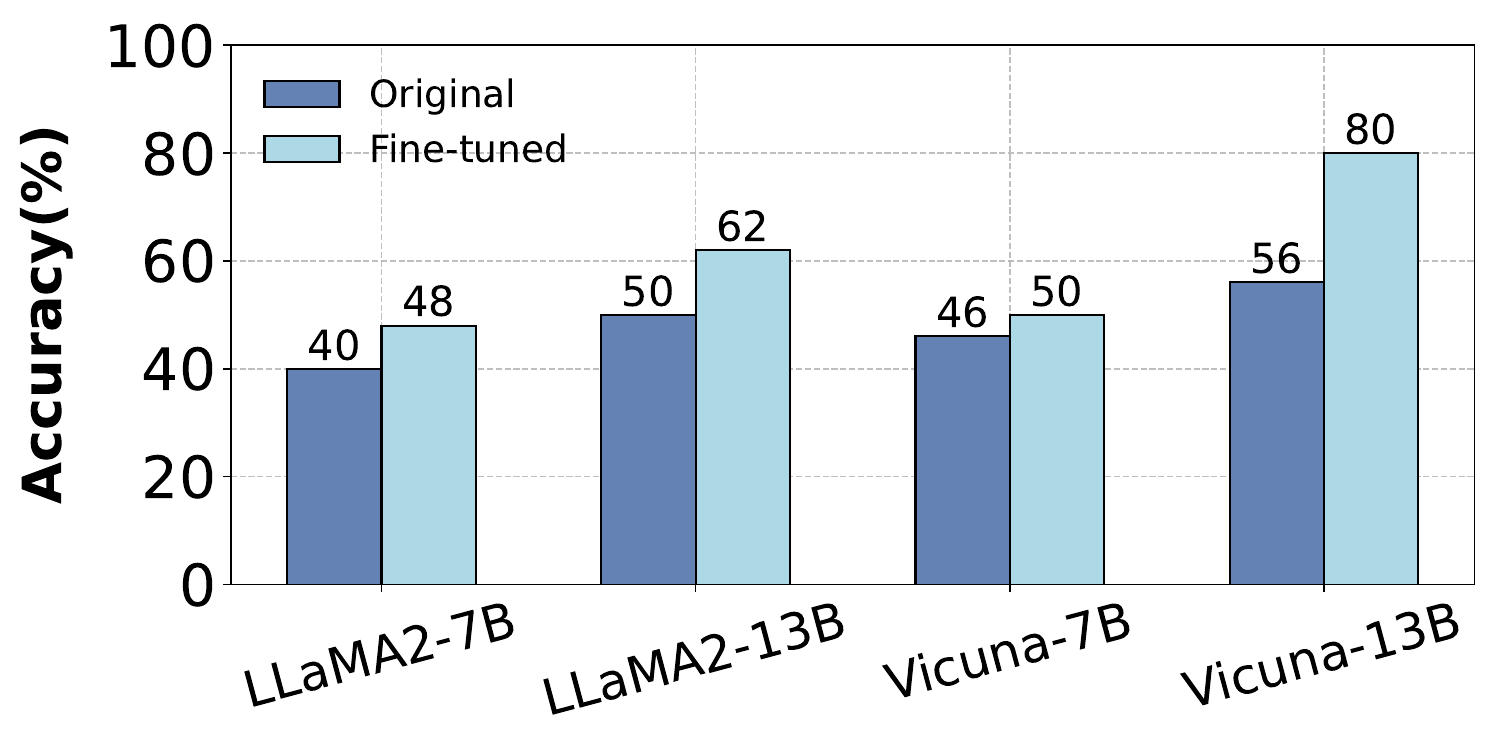}
    \vspace{-0.2cm}
    \caption{LLMs' Performance on Task 5 without fine-tuning (denoted as Original) or after being fine-tuned with training data of Task 1--4. 
    %We fine-tuned LLMs with the data of Tasks 1-4 and tested their performance on Task5, comparing the results with original LLMs. The samples tested before and after fine-tuning remained unchanged.
    }
    \label{fig:finetune2}
    \vspace{-0.2cm}
\end{figure}

\paragraph{Cross-task Learning Performance}
Compared with Task 1--4, Task 5 belongs to the higher cognition dimension \textbf{HOW}, and is more difficult for LLMs since it requires to generate a sentence satisfying the demand. An interesting research question is that, whether LLMs can well achieve Task 5 after learning the previous four tasks?
To answer this question, for each of Task 1--4 we sampled 60 instances from its training data, and mixed them with the equal amount (240) of general conversation instances from \href{https://huggingface.co/datasets/lmsys/lmsys-chat-1m}{lmsys-chat-1m} 
to fine-tune LLMs, which was used to guarantee LLMs' generative ability. Then, we evaluated the fine-tuned LLMs' performance on Task 5, of which the performance is depicted in Figure~\ref{fig:finetune2}. As well, LLMs' performance without fine-tuning, denoted as Original, is also displayed in the figure. 
The results indicate that, all the tested LLMs indeed enhanced their LFU performance through learning Task 1--4, also justifying their good cross-task learning capability of LFU tasks.

\section{Conclusion}
To evaluate LLMs' LFU performance, we propose five concrete tasks from three cognition dimensions WHAT, WHY, and HOW. Towards these tasks, we constructed a high quality dataset LFUD, which has been proven helpful by our extensive experiments to enhance LLMs' capability of logical reasoning.
We hope our work in this paper is instructive and our LFUD becomes a valuable resource for further research on LFU.

\section*{Limitations}

Although we argue that enhancing LLMs' logical reasoning capability through enabling LLMs to understand logical fallacies is language-independent, we should still acknowledge that the data and experiments of our work were only in English. As we know, LLMs might have different performance on many tasks including logical reasoning, across different languages. Therefore, the effectiveness of our solution proposed in this paper may vary when applied to other languages.

\section*{Ethical Considerations}
%All authors of this work abide by the provided Code of Ethics. The quality of manual proofreading for logical fallacy sentences is ensured through a double-check strategy outlined in Appendix \ref{sec:appendix_crowdsourcing}. We ensure that the privacy rights of all members for proofreading are respected in the process. 
% At first, no sensitive information of real-world people exists in the propositions and the sentences generated by GPT-4 in our LFUD. 
At first, all authors of this work abide by the provided Code of Ethics. The quality of manual proofreading for logical fallacy sentences is ensured through a double-check strategy outlined in Appendix \ref{sec:appendix_crowdsourcing}. We ensure that the privacy rights of all members for proofreading are respected in the process. 
Besides, synthetic data generated by LLMs may involve potential ethical risks regarding fairness and bias \cite{bommasani2021opportunities, blodgett-etal-2020-language}, which results in further consideration when they are employed in downstream tasks. Although our dataset LFUD was built for better understanding logical fallacies, which is not intended for safety-critical applications, we still asked our members for proofreading to refine the offensive and harmful data generated by GPT-4. Despite these considerations, there may still be some unsatisfactory data that goes unnoticed in our final dataset.

\section*{Acknowledgements}

This work was supported by Chinese
NSF Major Research Plan (No. 92270121), Youth Fund (No. 62102095), Shanghai Science and Technology Innovation Action Plan (No. 21511100401). The computations in this research were performed using the CFFF platform of Fudan University.
% This document has been adapted
% by Steven Bethard, Ryan Cotterell and Rui Yan
% from the instructions for earlier ACL and NAACL proceedings, including those for 
% ACL 2019 by Douwe Kiela and Ivan Vuli\'{c},
% NAACL 2019 by Stephanie Lukin and Alla Roskovskaya, 
% ACL 2018 by Shay Cohen, Kevin Gimpel, and Wei Lu, 
% NAACL 2018 by Margaret Mitchell and Stephanie Lukin,
% Bib\TeX{} suggestions for (NA)ACL 2017/2018 from Jason Eisner,
% ACL 2017 by Dan Gildea and Min-Yen Kan, 
% NAACL 2017 by Margaret Mitchell, 
% ACL 2012 by Maggie Li and Michael White, 
% ACL 2010 by Jing-Shin Chang and Philipp Koehn, 
% ACL 2008 by Johanna D. Moore, Simone Teufel, James Allan, and Sadaoki Furui, 
% ACL 2005 by Hwee Tou Ng and Kemal Oflazer, 
% ACL 2002 by Eugene Charniak and Dekang Lin, 
% and earlier ACL and EACL formats written by several people, including
% John Chen, Henry S. Thompson and Donald Walker.
% Additional elements were taken from the formatting instructions of the \emph{International Joint Conference on Artificial Intelligence} and the \emph{Conference on Computer Vision and Pattern Recognition}.

% \bibliographystyle{acl_natbib}
\bibliography{acl_latex}

\begin{thebibliography}{48}
\expandafter\ifx\csname natexlab\endcsname\relax\def\natexlab#1{#1}\fi

\bibitem[{Abaskohi et~al.(2022)Abaskohi, Rasouli, Zeraati, and Bahrak}]{abaskohi-etal-2022-utnlp}
Amirhossein Abaskohi, Arash Rasouli, Tanin Zeraati, and Behnam Bahrak. 2022.
\newblock \href {https://doi.org/10.18653/v1/2022.semeval-1.135} {{UTNLP} at {S}em{E}val-2022 task 6: A comparative analysis of sarcasm detection using generative-based and mutation-based data augmentation}.
\newblock In \emph{Proceedings of the 16th International Workshop on Semantic Evaluation (SemEval-2022)}, pages 962--969, Seattle, United States. Association for Computational Linguistics.

\bibitem[{An et~al.(2023)An, Ma, Lin, Zheng, Lou, and Chen}]{an2023learning}
Shengnan An, Zexiong Ma, Zeqi Lin, Nanning Zheng, Jian-Guang Lou, and Weizhu Chen. 2023.
\newblock Learning from mistakes makes llm better reasoner.
\newblock \emph{arXiv preprint arXiv:2310.20689}.

\bibitem[{Aristotle(2006)}]{aristotle2006sophistical}
Aristotle. 2006.
\newblock \emph{On sophistical refutations}.
\newblock ReadHowYouWant. com.

\bibitem[{bench authors(2023)}]{srivastava2023beyond}
BIG bench authors. 2023.
\newblock \href {https://openreview.net/forum?id=uyTL5Bvosj} {Beyond the imitation game: Quantifying and extrapolating the capabilities of language models}.
\newblock \emph{Transactions on Machine Learning Research}.

\bibitem[{Blair-Stanek et~al.(2023)Blair-Stanek, Holzenberger, and Van~Durme}]{blair2023can}
Andrew Blair-Stanek, Nils Holzenberger, and Benjamin Van~Durme. 2023.
\newblock Can gpt-3 perform statutory reasoning?
\newblock \emph{arXiv preprint arXiv:2302.06100}.

\bibitem[{Blodgett et~al.(2020)Blodgett, Barocas, Daum{\'e}~III, and Wallach}]{blodgett-etal-2020-language}
Su~Lin Blodgett, Solon Barocas, Hal Daum{\'e}~III, and Hanna Wallach. 2020.
\newblock \href {https://doi.org/10.18653/v1/2020.acl-main.485} {Language (technology) is power: A critical survey of {``}bias{''} in {NLP}}.
\newblock In \emph{Proceedings of the 58th Annual Meeting of the Association for Computational Linguistics}, pages 5454--5476, Online. Association for Computational Linguistics.

\bibitem[{Bommasani et~al.(2021)Bommasani, Hudson, Adeli, Altman, Arora, von Arx, Bernstein, Bohg, Bosselut, Brunskill et~al.}]{bommasani2021opportunities}
Rishi Bommasani, Drew~A Hudson, Ehsan Adeli, Russ Altman, Simran Arora, Sydney von Arx, Michael~S Bernstein, Jeannette Bohg, Antoine Bosselut, Emma Brunskill, et~al. 2021.
\newblock On the opportunities and risks of foundation models.
\newblock \emph{arXiv preprint arXiv:2108.07258}.

\bibitem[{Chen et~al.(2023{\natexlab{a}})Chen, Wang, Yang, Han, Hong, Mi, Xu, Liu, Huang, Li et~al.}]{chen2023gaining}
Kai Chen, Chunwei Wang, Kuo Yang, Jianhua Han, Lanqing Hong, Fei Mi, Hang Xu, Zhengying Liu, Wenyong Huang, Zhenguo Li, et~al. 2023{\natexlab{a}}.
\newblock Gaining wisdom from setbacks: Aligning large language models via mistake analysis.
\newblock \emph{arXiv preprint arXiv:2310.10477}.

\bibitem[{Chen et~al.(2023{\natexlab{b}})Chen, Ma, Song, Cao, Zhang, and Li}]{chen2023learning}
Meiqi Chen, Yubo Ma, Kaitao Song, Yixin Cao, Yan Zhang, and Dongsheng Li. 2023{\natexlab{b}}.
\newblock \href {http://arxiv.org/abs/2310.09158} {Learning to teach large language models logical reasoning}.

\bibitem[{Chiang et~al.(2023)Chiang, Li, Lin, Sheng, Wu, Zhang, Zheng, Zhuang, Zhuang, Gonzalez et~al.}]{chiang2023vicuna}
Wei-Lin Chiang, Zhuohan Li, Zi~Lin, Ying Sheng, Zhanghao Wu, Hao Zhang, Lianmin Zheng, Siyuan Zhuang, Yonghao Zhuang, Joseph~E Gonzalez, et~al. 2023.
\newblock Vicuna: An open-source chatbot impressing gpt-4 with 90\%* chatgpt quality.
\newblock \emph{See https://vicuna. lmsys. org (accessed 14 April 2023)}.

\bibitem[{Chung et~al.(2023)Chung, Kamar, and Amershi}]{chung2023increasing}
John Joon~Young Chung, Ece Kamar, and Saleema Amershi. 2023.
\newblock Increasing diversity while maintaining accuracy: Text data generation with large language models and human interventions.
\newblock \emph{arXiv preprint arXiv:2306.04140}.

\bibitem[{Cresswell(1973)}]{Cresswell1973-CRELAL-3}
Maxwell~John Cresswell. 1973.
\newblock \emph{Logics and Languages}.
\newblock Routledge, London, England.

\bibitem[{Eldan and Li(2023)}]{eldan2023tinystories}
Ronen Eldan and Yuanzhi Li. 2023.
\newblock Tinystories: How small can language models be and still speak coherent english?
\newblock \emph{arXiv preprint arXiv:2305.07759}.

\bibitem[{Goffredo et~al.(2022)Goffredo, Haddadan, Vorakitphan, Cabrio, and Villata}]{goffredo2022fallacious}
Pierpaolo Goffredo, Shohreh Haddadan, Vorakit Vorakitphan, Elena Cabrio, and Serena Villata. 2022.
\newblock Fallacious argument classification in political debates.
\newblock In \emph{Proceedings of the Thirty-First International Joint Conference on Artificial Intelligence, IJCAI}, pages 4143--4149.

\bibitem[{Habernal et~al.(2018)Habernal, Wachsmuth, Gurevych, and Stein}]{habernal-etal-2018-name}
Ivan Habernal, Henning Wachsmuth, Iryna Gurevych, and Benno Stein. 2018.
\newblock \href {https://doi.org/10.18653/v1/N18-1036} {Before name-calling: Dynamics and triggers of ad hominem fallacies in web argumentation}.
\newblock In \emph{Proceedings of the 2018 Conference of the North {A}merican Chapter of the Association for Computational Linguistics: Human Language Technologies, Volume 1 (Long Papers)}, pages 386--396, New Orleans, Louisiana. Association for Computational Linguistics.

\bibitem[{Han et~al.(2022)Han, Schoelkopf, Zhao, Qi, Riddell, Benson, Sun, Zubova, Qiao, Burtell et~al.}]{han2022folio}
Simeng Han, Hailey Schoelkopf, Yilun Zhao, Zhenting Qi, Martin Riddell, Luke Benson, Lucy Sun, Ekaterina Zubova, Yujie Qiao, Matthew Burtell, et~al. 2022.
\newblock Folio: Natural language reasoning with first-order logic.
\newblock \emph{arXiv preprint arXiv:2209.00840}.

\bibitem[{Hausman(2012)}]{Hausman2012-HAULAP}
Alan Hausman. 2012.
\newblock \emph{Logic and Philosophy: A Modern Introduction}.
\newblock Wadsworth, Cengage Learning, Boston, MA.

\bibitem[{Huang and Chang(2022)}]{huang2022towards}
Jie Huang and Kevin Chen-Chuan Chang. 2022.
\newblock Towards reasoning in large language models: A survey.
\newblock \emph{arXiv preprint arXiv:2212.10403}.

\bibitem[{Hurley(2000)}]{Hurley2000-HURACI}
Patrick~J. Hurley. 2000.
\newblock \emph{A Concise Introduction to Logic}.
\newblock Wadsworth, Belmont, CA.

\bibitem[{Iwa{\'n}ska(1993)}]{iwanska1993logical}
Lucja Iwa{\'n}ska. 1993.
\newblock Logical reasoning in natural language: It is all about knowledge.
\newblock \emph{Minds and Machines}, 3:475--510.

\bibitem[{Jin et~al.(2022)Jin, Lalwani, Vaidhya, Shen, Ding, Lyu, Sachan, Mihalcea, and Schoelkopf}]{jin-etal-2022-logical}
Zhijing Jin, Abhinav Lalwani, Tejas Vaidhya, Xiaoyu Shen, Yiwen Ding, Zhiheng Lyu, Mrinmaya Sachan, Rada Mihalcea, and Bernhard Schoelkopf. 2022.
\newblock \href {https://doi.org/10.18653/v1/2022.findings-emnlp.532} {Logical fallacy detection}.
\newblock In \emph{Findings of the Association for Computational Linguistics: EMNLP 2022}, pages 7180--7198, Abu Dhabi, United Arab Emirates. Association for Computational Linguistics.

\bibitem[{Joshi et~al.(2020)Joshi, Aditya, Sathe, and Choudhury}]{joshi2020taxinli}
Pratik Joshi, Somak Aditya, Aalok Sathe, and Monojit Choudhury. 2020.
\newblock Taxinli: Taking a ride up the nlu hill.
\newblock \emph{arXiv preprint arXiv:2009.14505}.

\bibitem[{Josifoski et~al.(2023)Josifoski, Sakota, Peyrard, and West}]{josifoski2023exploiting}
Martin Josifoski, Marija Sakota, Maxime Peyrard, and Robert West. 2023.
\newblock Exploiting asymmetry for synthetic training data generation: Synthie and the case of information extraction.
\newblock \emph{arXiv preprint arXiv:2303.04132}.

\bibitem[{Kowalski(1974)}]{kowalski1974logic}
Robert Kowalski. 1974.
\newblock \emph{Logic for problem solving}.
\newblock Department of Computational Logic, Edinburgh University.

\bibitem[{Li et~al.(2023)Li, Zhu, Lu, and Yin}]{li2023synthetic}
Zhuoyan Li, Hangxiao Zhu, Zhuoran Lu, and Ming Yin. 2023.
\newblock Synthetic data generation with large language models for text classification: Potential and limitations.
\newblock \emph{arXiv preprint arXiv:2310.07849}.

\bibitem[{Liu et~al.(2021)Liu, Cui, Liu, and Zhang}]{liu2021natural}
Hanmeng Liu, Leyang Cui, Jian Liu, and Yue Zhang. 2021.
\newblock Natural language inference in context-investigating contextual reasoning over long texts.
\newblock In \emph{Proceedings of the AAAI Conference on Artificial Intelligence}, volume~35, pages 13388--13396.

\bibitem[{Liu et~al.(2020)Liu, Cui, Liu, Huang, Wang, and Zhang}]{liu2020logiqa}
Jian Liu, Leyang Cui, Hanmeng Liu, Dandan Huang, Yile Wang, and Yue Zhang. 2020.
\newblock Logiqa: A challenge dataset for machine reading comprehension with logical reasoning.
\newblock \emph{arXiv preprint arXiv:2007.08124}.

\bibitem[{Maqsud(2015)}]{maqsud2015synthetic}
Umar Maqsud. 2015.
\newblock Synthetic text generation for sentiment analysis.
\newblock In \emph{Proceedings of the 6th Workshop on Computational Approaches to Subjectivity, Sentiment and Social Media Analysis}, pages 156--161.

\bibitem[{Martino et~al.(2020)Martino, Barr{\'o}n-Cedeno, Wachsmuth, Petrov, and Nakov}]{martino2020semeval}
G~Martino, Alberto Barr{\'o}n-Cedeno, Henning Wachsmuth, Rostislav Petrov, and Preslav Nakov. 2020.
\newblock Semeval-2020 task 11: Detection of propaganda techniques in news articles.
\newblock \emph{arXiv preprint arXiv:2009.02696}.

\bibitem[{Mitra et~al.(2023)Mitra, Corro, Mahajan, Codas, Simoes, Agrawal, Chen, Razdaibiedina, Jones, Aggarwal, Palangi, Zheng, Rosset, Khanpour, and Awadallah}]{mitra2023orca}
Arindam Mitra, Luciano~Del Corro, Shweti Mahajan, Andres Codas, Clarisse Simoes, Sahaj Agrawal, Xuxi Chen, Anastasia Razdaibiedina, Erik Jones, Kriti Aggarwal, Hamid Palangi, Guoqing Zheng, Corby Rosset, Hamed Khanpour, and Ahmed Awadallah. 2023.
\newblock \href {http://arxiv.org/abs/2311.11045} {Orca 2: Teaching small language models how to reason}.

\bibitem[{M{\o}ller et~al.(2023)M{\o}ller, Dalsgaard, Pera, and Aiello}]{moller2023prompt}
Anders~Giovanni M{\o}ller, Jacob~Aarup Dalsgaard, Arianna Pera, and Luca~Maria Aiello. 2023.
\newblock Is a prompt and a few samples all you need? using gpt-4 for data augmentation in low-resource classification tasks.
\newblock \emph{arXiv preprint arXiv:2304.13861}.

\bibitem[{Ontanon et~al.(2022)Ontanon, Ainslie, Cvicek, and Fisher}]{ontanon2022logicinference}
Santiago Ontanon, Joshua Ainslie, Vaclav Cvicek, and Zachary Fisher. 2022.
\newblock Logicinference: A new dataset for teaching logical inference to seq2seq models.
\newblock \emph{arXiv preprint arXiv:2203.15099}.

\bibitem[{OpenAI(2023)}]{openai2023gpt}
R~OpenAI. 2023.
\newblock Gpt-4 technical report. arxiv 2303.08774.
\newblock \emph{View in Article}.

\bibitem[{Ouyang et~al.(2022)Ouyang, Wu, Jiang, Almeida, Wainwright, Mishkin, Zhang, Agarwal, Slama, Ray et~al.}]{ouyang2022training}
Long Ouyang, Jeffrey Wu, Xu~Jiang, Diogo Almeida, Carroll Wainwright, Pamela Mishkin, Chong Zhang, Sandhini Agarwal, Katarina Slama, Alex Ray, et~al. 2022.
\newblock Training language models to follow instructions with human feedback.
\newblock \emph{Advances in Neural Information Processing Systems}, 35:27730--27744.

\bibitem[{Papanikolaou and Pierleoni(2020)}]{papanikolaou2020dare}
Yannis Papanikolaou and Andrea Pierleoni. 2020.
\newblock Dare: Data augmented relation extraction with gpt-2.
\newblock \emph{arXiv preprint arXiv:2004.13845}.

\bibitem[{Payandeh et~al.(2023)Payandeh, Pluth, Hosier, Xiao, and Gurbani}]{payandeh2023susceptible}
Amirreza Payandeh, Dan Pluth, Jordan Hosier, Xuesu Xiao, and Vijay~K Gurbani. 2023.
\newblock How susceptible are llms to logical fallacies?
\newblock \emph{arXiv preprint arXiv:2308.09853}.

\bibitem[{Sennrich et~al.(2015)Sennrich, Haddow, and Birch}]{sennrich2015improving}
Rico Sennrich, Barry Haddow, and Alexandra Birch. 2015.
\newblock Improving neural machine translation models with monolingual data.
\newblock \emph{arXiv preprint arXiv:1511.06709}.

\bibitem[{Sourati et~al.(2023)Sourati, Ilievski, Sandlin, and Mermoud}]{sourati2023case}
Zhivar Sourati, Filip Ilievski, H{\^o}ng-{\^A}n Sandlin, and Alain Mermoud. 2023.
\newblock Case-based reasoning with language models for classification of logical fallacies.
\newblock \emph{arXiv preprint arXiv:2301.11879}.

\bibitem[{Stab and Gurevych(2017)}]{stab-gurevych-2017-recognizing}
Christian Stab and Iryna Gurevych. 2017.
\newblock \href {https://aclanthology.org/E17-1092} {Recognizing insufficiently supported arguments in argumentative essays}.
\newblock In \emph{Proceedings of the 15th Conference of the {E}uropean Chapter of the Association for Computational Linguistics: Volume 1, Long Papers}, pages 980--990, Valencia, Spain. Association for Computational Linguistics.

\bibitem[{Swanborn(2010)}]{swanborn2010case}
Peter Swanborn. 2010.
\newblock Case study research: What, why and how?
\newblock \emph{Case study research}, pages 1--192.

\bibitem[{Teng et~al.(2023)Teng, Ning, Liu, Zhou, Zhang et~al.}]{teng2023glore}
Zhiyang Teng, Ruoxi Ning, Jian Liu, Qiji Zhou, Yue Zhang, et~al. 2023.
\newblock Glore: Evaluating logical reasoning of large language models.
\newblock \emph{arXiv preprint arXiv:2310.09107}.

\bibitem[{Tindale(2007)}]{tindale2007fallacies}
Christopher~W Tindale. 2007.
\newblock \emph{Fallacies and argument appraisal}.
\newblock Cambridge University Press.

\bibitem[{Touvron et~al.(2023)Touvron, Lavril, Izacard, Martinet, Lachaux, Lacroix, Rozi{\`e}re, Goyal, Hambro, Azhar et~al.}]{touvron2023llama}
Hugo Touvron, Thibaut Lavril, Gautier Izacard, Xavier Martinet, Marie-Anne Lachaux, Timoth{\'e}e Lacroix, Baptiste Rozi{\`e}re, Naman Goyal, Eric Hambro, Faisal Azhar, et~al. 2023.
\newblock Llama: Open and efficient foundation language models.
\newblock \emph{arXiv preprint arXiv:2302.13971}.

\bibitem[{Wang et~al.(2022)Wang, Liu, Zhong, Zhou, Wei, Chen, and Duan}]{wang2022lsat}
Siyuan Wang, Zhongkun Liu, Wanjun Zhong, Ming Zhou, Zhongyu Wei, Zhumin Chen, and Nan Duan. 2022.
\newblock From lsat: The progress and challenges of complex reasoning.
\newblock \emph{IEEE/ACM Transactions on Audio, Speech, and Language Processing}, 30:2201--2216.

\bibitem[{Yanaka et~al.(2019)Yanaka, Mineshima, Bekki, Inui, Sekine, Abzianidze, and Bos}]{yanaka2019help}
Hitomi Yanaka, Koji Mineshima, Daisuke Bekki, Kentaro Inui, Satoshi Sekine, Lasha Abzianidze, and Johan Bos. 2019.
\newblock Help: A dataset for identifying shortcomings of neural models in monotonicity reasoning.
\newblock \emph{arXiv preprint arXiv:1904.12166}.

\bibitem[{Yu et~al.(2023)Yu, Zhang, and Wang}]{yu2023nature}
Fei Yu, Hongbo Zhang, and Benyou Wang. 2023.
\newblock Nature language reasoning, a survey.
\newblock \emph{arXiv preprint arXiv:2303.14725}.

\bibitem[{Yu et~al.(2020)Yu, Jiang, Dong, and Feng}]{yu2020reclor}
Weihao Yu, Zihang Jiang, Yanfei Dong, and Jiashi Feng. 2020.
\newblock Reclor: A reading comprehension dataset requiring logical reasoning.
\newblock \emph{arXiv preprint arXiv:2002.04326}.

\bibitem[{Zhang et~al.(2023)Zhang, Huang, Li, Naik, and Xing}]{zhang2023improved}
Hanlin Zhang, Jiani Huang, Ziyang Li, Mayur Naik, and Eric Xing. 2023.
\newblock Improved logical reasoning of language models via differentiable symbolic programming.
\newblock \emph{arXiv preprint arXiv:2305.03742}.

\end{thebibliography}
\appendix

\begin{table*}
\centering
\resizebox{\linewidth}{!}{
\begin{tabular}{lp{8cm}p{7cm}}
\hline
\textbf{Fallacy Type} & \textbf{Description} & \textbf{Example} \\
\hline
\textbf{Faulty Generalization} & Faulty generalization occurs when a conclusion about all or many instances of a phenomenon is drawn from one or a few instances of that phenomenon. & Kevin, who is a teenager, enjoys playing chess. Therefore, all teenagers must enjoy playing chess.\\
\hline
\textbf{False Causality} & False causality occurs when an argument jumps to a conclusion implying a causal relationship without supporting evidence. & Whenever David goes hiking in the mountains, it's a sunny day. Clearly, David's hiking trips cause sunny weather. \\
\hline
\textbf{Circular Claim} & Circular reasoning occurs when an argument uses the claim it is trying to prove as proof that the claim is true. & Some students are not serious about their studies because they do not focus on their studies.  \\
\hline
\textbf{Ad Populum} & Ad populum occurs when an argument is based on affirming that something is real or better because the majority thinks so. & It's widely believed that Nancy relocated to another city, so it must be true. \\
\hline
\textbf{Ad Hominem} & Ad hominem is an irrelevant attack towards the person or some aspect of the person who is making the argument, instead of addressing the argument or position directly. & John claims that all people should obey the rules of the road.  But John has received several speeding tickets in the past.  Therefore, it's not necessary to obey the rules of the road. \\
\hline
\textbf{Deductive Fallacy} & Deductive fallacy occurs when there is a logical flaw in the reasoning behind the argument, such as Affirming the consequent, Denying the antecedent, Affirming a disjunct and so on. & Should Lucy feel alone, she will surely adopt a puppy.  It's evident Lucy has adopted a puppy.  Therefore, it must be that Lucy is feeling lonely. \\
\hline
\textbf{Appeal to Emotion} & Appeal to emotion is when emotion is used in place of reason to support an argument in place of reason, such as pity, fear, anger, etc. & Jack had his wallet stolen at the concert, think about how desperate and helpless Jack is now, how can we not help him? \\
\hline
\textbf{False Dilemma} & False dilemma occurs when incorrect limitations are made on the possible options in a scenario when there could be other options. & Most museums will be closed on Mondays either due to low visitor turnout, or due to their disregard for public interest. \\
\hline
\textbf{Equivocation} & Equivocation is an argument which uses a key term or phrase in an ambiguous way, with one meaning in one portion of the argument and then another meaning in another portion of the argument. & All stars are exploding balls of gas.  Miley Cyrus is a star.  Therefore, Miley Cyrus is an exploding ball of gas. \\
\hline
\textbf{Fallacy of Extension} & Fallacy of extension is an argument that attacks an exaggerated or caricatured version of your opponent's position. & Alex: All flowers don't stay open forever.  Jamie: So you're saying that all flowers die instantly after they bloom? \\
\hline
\textbf{Fallacy of Relevance} & Fallacy of relevance, which is also known as Red Herring, occurs when the speaker attempts to divert attention from the primary argument by offering a point that does not suffice as counterpoint/supporting evidence (even if it is true). & A portion of the inhabitants of this city have a fever, but have you considered the high unemployment rate? \\
\hline
\textbf{Fallacy of Credibility} & Fallacy of credibility is when an appeal is made to some form of ethics, authority, or credibility. & Sharon, an acclaimed pianist with years of experience, claims that practicing every day will increase your piano skills by 50\%.  She's an expert, therefore we should believe her. \\
\hline
\textbf{Intentional Fallacy} & Intentional fallacy is a custom category for when an argument has some element that shows the intent of a speaker to win an argument without actual supporting evidence. & Since no one can prove that Peter didn't come to China last year, he must have.\\
\hline
\end{tabular}}
\caption{Descriptions and examples of 13 logical fallacy types}
\label{tab:fallacy}
\end{table*}

\section{Details of Five LFU Tasks}
\label{sec:appendix_taskintro}
We list the definitions and examples of our five tasks below.
\newline

\noindent \textbf{Dimension: WHAT}
\begin{itemize}
    \setlength{\itemsep}{0pt} % 定义列表项之间的距离
    \setlength{\parsep}{0pt}  % 定义段落之间的距离
    \setlength{\parskip}{0pt} % 定义列表和其他元素之间的距离
    \item \textbf{Task1:} Identification
    % \addtolength{\itemindent}{-2em} 
    \item \textbf{Definition:} Identify whether the given sentence has logical fallacy. 
    % \addtolength{\itemindent}{2em}
    \item \textbf{Example: }\newline Sentence: Many people believe most museums will be closed on Mondays, therefore it's a fact. \newline Identify if there is any logical fallacy in the sentence. \newline A) Yes, there is a logical fallacy. \newline B) No, there is no logical fallacy.
\end{itemize}
% \vspace{-\baselineskip} % 移除当前行的垂直空间

% \paragraph{}
\begin{itemize}
    \setlength{\itemsep}{0pt} % 定义列表项之间的距离
    \setlength{\parsep}{0pt}  % 定义段落之间的距离
    \setlength{\parskip}{0pt} % 定义列表和其他元素之间的距离
    \item \textbf{Task2:} Classification
    \item \textbf{Definition:} Select the sentence belonging to a certain type of logical fallacy.
    \item \textbf{Example: }\newline Circular reasoning occurs when an argument uses the claim it is trying to prove as proof that the claim is true. \newline Select which among the following options demonstrates the logical fallacy of circular reasoning. \newline A) Most people believe that Rebecca doesn't like spicy food, therefore it must be true. \newline B) Rebecca, a renowned food critic, does not like spicy food. Hence, spicy food is not good. \newline C) Rebecca either refrains from spicy food due to discomfort it causes her, or she lacks well-developed taste buds. \newline D) Rebecca doesn't like spicy food because she dislikes spicy food. 
    
% \uline{D) Rebecca doesn't like spicy food because she dislikes spicy food.}
\end{itemize}

\paragraph{Dimension: WHY}
\begin{itemize}
    \setlength{\itemsep}{0pt} % 定义列表项之间的距离
    \setlength{\parsep}{0pt}  % 定义段落之间的距离
    \setlength{\parskip}{0pt} % 定义列表和其他元素之间的距离
    \item \textbf{Task3:} Deduction
    \item \textbf{Definition:} Derive the conclusion from the premise according to a certain type of logical fallacy.
    \item \textbf{Example: }\newline Faulty generalization occurs when a conclusion about all or many instances of a phenomenon is drawn from one or a few instances of that phenomenon. \newline The premise is known: Bob painted his house green and he is a homeowner. \newline With which of the two conclusions can the premise be coupled to create logical fallacy of faulty generalization? \newline A) Green is the most popular house color. \newline B) All homeowners paint their houses green.
\end{itemize}

% \paragraph{}
\begin{itemize}
    \setlength{\itemsep}{0pt} % 定义列表项之间的距离
    \setlength{\parsep}{0pt}  % 定义段落之间的距离
    \setlength{\parskip}{0pt} % 定义列表和其他元素之间的距离
    \item \textbf{Task4:} Backward Deduction
    \item \textbf{Definition:} Infer the premise from the conclusion according to a certain type of logical fallacy.
    \item \textbf{Example: }\newline Ad populum occurs when an argument is based on affirming that something is real or better because the majority thinks so. \newline The conclusion is known: Cynthia's painting must be a masterpiece. \newline With which of the two premises can the conclusion be coupled to create the logical fallacy of ad populum?\newline A) People widely agree that Cynthia made a beautiful painting. \newline B) A famous art critic praised Cynthia's painting.
\end{itemize}

\paragraph{Dimension: HOW}
\begin{itemize}
    \setlength{\itemsep}{0pt} % 定义列表项之间的距离
    \setlength{\parsep}{0pt}  % 定义段落之间的距离
    \setlength{\parskip}{0pt} % 定义列表和其他元素之间的距离
    \item \textbf{Task5:} Modification
    \item \textbf{Definition:} Correct the logical fallacy in the given sentence.
    \item \textbf{Example: }\newline Original sentence: Person A: The garden needs watering. Person B: So you're saying we should neglect everything else and just focus on the garden? \newline Correct the logical fallacy in the original sentence and output the modified sentence without any logical fallacy.
% Modified sentence:
\end{itemize}

\section{Details of Logic Fallacy Types}\label{sec:appendix_fallacy_type}
In Table \ref{tab:fallacy}, we showcase the description and examples of 13 logical fallacy types.

\section{Details of Manual Proofreading}
\label{sec:appendix_crowdsourcing}
% 满足条件的逻辑谬误语句是设计任务的基础。尽管GPT-4拥有强大的通用能力，但为了保证严谨性，我们还是进行了人工校对。在这个过程中，我们组建了一个精通语言规则和逻辑学的审核团队，团队由4人组成，包括1名逻辑学家和3名研究生，分别从事语言学、逻辑学和计算机研究，都能够理解并且区分不同类型的逻辑谬误。评估标准为句子是否满足相应类别的逻辑谬误，每个句子将会被分配给两位同学，如有相同意见则确认无误，如有意见分歧则请另一位同学判断。若三人意见无法达成一致，则请逻辑学家来做最后定夺。每个句子都会被仔细检查，任何错误、逻辑不一致或需要改进的地方都会被记录下来，并由人工重新措辞。
% 评估标准将严格分为两大类--结构完整性和逻辑一致性。结构完整性将侧重于语法的正确性、标点符号的准确性以及句法的正确使用。另一方面，逻辑一致性将确保消除句子中的矛盾语句、歧义和任何形式的逻辑谬误。这还将确保语句在特定的语境或主题下是合理的。
% 为了提高这一过程的效率，每个句子最初都将通过 Grammarly 或类似的语法检查工具进行处理，以剔除基本的语法和词汇错误。之后，专家团队将进行人工审核，同时考虑到可能存在的额外语境和细微语言。
The evaluation standard is strictly classified into two main categories: structural integrity and validity of fallacies. Structural integrity focuses on the correctness of grammar, the accuracy of punctuation, and the proper use of syntax. On the other hand, the validity of fallacies ensures that, under specific contexts or themes, the sentences satisfy the need for specific type of logical fallacy. Besides, any offensive and harmful data will be refined during the proofreading.

In this process, we assembled an expert team proficient in linguistics and logic. This team comprises four members, including one logician and three graduate students, who are engaged in linguistics, logic, and computer science respectively. They each have the ability to understand and classify various types of logical fallacies.

To enhance the efficiency of this process, each sentence was initially processed through Grammarly, eliminating basic grammatical and lexical errors. Subsequently, our expert team manually reviewed the content. Each sentence was assigned to two team members for review. A consensus confirmed the sentence met the requirements, but in case of disagreement, the third team member would be consulted. If three members cannot achieve consensus, the logician will make the final decision.

\section{Examples of Logical Reasoning Datasets}
\label{sec:appendix_logicalreasongdataset}
We illustrate data examples of four logical reasoning datasets selected in our experiments, including FOLIO, TaxiNLI, LogiQA, and Reclor. 
\newline

\noindent \textbf{FOLIO}

\textbf{Premise}: Beasts of Prey is either a fantasy novel or a science fiction novel.  Science fiction novels are not about mythological creatures. Beasts of Prey Is about a creature known as the Shetani. Shetanis are mythological.

\textbf{Conclusion}: Beasts of prey isn’t a science fiction novel.

\textbf{Answer}: True
\newline

% \paragraph{TaxiNLI}\mbox{}\\
\noindent \textbf{TaxiNLI}

\textbf{Premise}: Even if auditors do not follow such other standards and methodologies, they may still serve as a useful source of guidance to auditors in planning their work under GAGAS.

\textbf{Hypothesis}: Auditors should ignore them when they follow other standards and methodologies.

\textbf{Label}: Contradiction
\newline

\noindent \textbf{LogiQA2.0}

\textbf{Passage}: For a television program about astrology, investigators went into the street and found twenty volunteers born under the sign of Gemini who were willing to be interviewed on the program and to take a personality test. The test confirmed the investigators' personal impressions that each of the volunteers was more sociable and extroverted than people are on average. This modest investigation thus supports the claim that one's astrological birth sign influences one's personality.

\textbf{Question}: Which one of the following, if true, indicates the most serious flaw in the method used by the investigators? 

A. People born under astrological signs other than Gemini have been judged by astrologers to be much less sociable than those born under Gemini.

B. There is not likely to be a greater proportion of people born under the sign of Gemini on the street than in the population as a whole.

C. People who are not sociable and extroverted are not likely to agree to participate in such an investigation.

D. The personal impressions the investigators first formed of other people have tended to be confirmed by the investigators' later experience of those people.

\textbf{Answer}: C
\newline

\noindent \textbf{Reclor}

\textbf{Context}: Geologist: A new method for forecasting earthquakes has reliably predicted several earthquakes. Unfortunately, this method can predict only that an earthquake will fall somewhere within a range of two and a half points on the Richter scale. Thus, since a difference of two and a half points can be the difference between a marginally perceptible shaking and a quake that causes considerable damage, the new method is unlikely to be useful.

\textbf{Question}: Which one of the following, if assumed, enables the geologist’s conclusion to be properly inferred?

A. An earthquake-forecasting method is unlikely to be useful unless its predictions always differentiate earthquakes that are barely noticeable from ones that result in substantial destruction.

B. Several well-established methods for forecasting earthquakes can predict within much narrower ranges than two and a half points on the Richter scale.

C. Even if an earthquake-forecasting method makes predictions within a very narrow range on the Richter scale, this method is not likely to be useful unless its predictions are reliable.

D. An earthquake-forecasting method has not been shown to be useful until it has been used to reliably predict a large number of earthquakes.

\textbf{Answer}: A

\end{document}